\documentclass[conference,10pt]{IEEEtran}
\usepackage{algpseudocode}
\usepackage[utf8]{inputenc} 
\usepackage[T1]{fontenc}    

\usepackage{booktabs}       
\usepackage{amsfonts}       
\usepackage{nicefrac}       
\usepackage{microtype}      
\usepackage{graphicx}

\usepackage{commath}
\usepackage{epic,epsfig}
\usepackage[ruled,vlined, linesnumbered]{algorithm2e}
\usepackage[draft]{hyperref}
\begin{document}
\title{PALMAR: Towards Adaptive Multi-inhabitant Activity Recognition in Point-Cloud Technology}
\author{Mohammad Arif Ul Alam, Md Mahmudur Rahman, Jared Q Widberg}

\author{\IEEEauthorblockN{Mohammad Arif Ul Alam, Md Mahmudur Rahman, Jared Q Widberg}
\IEEEauthorblockA{{Department of Computer Science, University of Massachusetts Lowell, MA, USA}
}}

\maketitle
\begin{abstract}
With the advancement of deep neural networks and computer vision-based Human Activity Recognition, employment of Point-Cloud Data technologies (LiDAR, mmWave) has seen a lot interests due to its privacy preserving nature. Given the high promise of accurate PCD technologies, we develop, PALMAR, a multiple-inhabitant activity recognition system by employing efficient signal processing and novel machine learning techniques to track individual person towards developing an adaptive multi-inhabitant tracking and HAR system. More specifically, we propose (i) a voxelized feature representation-based real-time PCD fine-tuning method, (ii) efficient clustering (DBSCAN and BIRCH), Adaptive Order Hidden Markov Model based multi-person tracking and crossover ambiguity reduction techniques and (iii) novel adaptive deep learning-based domain adaptation technique to improve the accuracy of HAR in presence of data scarcity and diversity (device, location and population diversity). We experimentally evaluate our framework and systems using (i) a real-time PCD collected by three devices (3D LiDAR and 79 GHz mmWave) from 6 participants, (ii) one publicly available 3D LiDAR activity data (28 participants) and (iii) an embedded hardware prototype system which provided promising HAR performances in multi-inhabitants (96\%) scenario with a 63\% improvement of multi-person tracking than state-of-art framework without losing significant system performances in the edge computing device.
\end{abstract}

\begin{IEEEkeywords}
Human Activity Recognition, Domain Adaptation, PCD Sensor Technology, Edge Computing
\end{IEEEkeywords}

\maketitle

\section{Introduction}
Though, recent advancement of internet-of-things (IoT) sensors and deep learning techniques result significant improvement in HAR, the most accurate Multiple Person Tracking and Human Activity Recognition (MPT-HAR) state-of-arts are still computer vision based techniques \cite{24, 27, 31, 40, 46,4,11,64,7,19,36,54,63,41,8}. Due to the lower acceptance of privacy-concerned camera-based frameworks, many researchers employed privacy-preserving PCD technologies such as Light Detection and Ranging (LiDAR), millimeter Wave (mmWave) and Ultrasound (US) to investigate MPT-HAR problem to reach the closest accuracy to camera-based techniques by introducing extensive signal processing and complex deep learning models \cite{24, 27, 31, 40}. In this paper, we propose a novel MPT-HAR system that can transfer the domain knowledge from accurate privacy concerned technology (say Computer Vision) to privacy-preserving PCD technologies by utilizing signal processing and adaptive machine learning techniques.

Many PCD technologies, such as Ultrasound, mmWave and LiDAR, have been used by researchers before for gait pattern identification \cite{benedek18}, activity recognition \cite{benedek18}, person tracking \cite{zhao19}. All of the above sensors provide PCD that are, in general, large data sets composed of 3D point data. PCD are derived from raw data scanned from physical objects such as building exteriors and interiors, humans, process plants, topographies, and manufactured items. More specifically for PCD generating systems that emit light pulses outside visible light spectrum and capture the duration of its return. The distance vector of returned pulse is saved as point of cloud which is represented as $x$, $y$ and $z$ values in 3D space.

Previously, expensive PCD systems have been used to solve different problems such as object detection \cite{rehder14}, distance identification \cite{died15}, 3D imaging \cite{died15}, multiple person tracking \cite{zhao19} and gait recognition \cite{zhao19}. The above solutions have been applied on many applications such as automated driving \cite{died15}, remote health monitoring \cite{zhao19} and elderly care \cite{zhao19}. In current state-of-art methods, employment of intense signal processing and deep learning techniques on PCD Data resulted maximum accuracy of multiple person tracking with 89\% \cite{zhao19} and multiple person HAR with 75\% \cite{benedek18} which needs significant improvement.

In this paper, we argue that privacy preserving PCD data based HAR can be significantly improved by existing computer vision data which needs careful utilization of  signal processing as well as novel domain adaptation techniques. In this regard, we propose novel trios: Adaptive Order Hidden Markov Model to track multiple persons, Crossover Path Disambiguation Algorithm (CPDA) to improve tracking and variational autoencoder-based domain adaptation algorithm to improve activity recognition. The core of our proposed architecture is an adaptive deep learning framework that consists of the following modules: \emph{Person tracker} module combines voxel representator, DBSCAN and BIRCH clustering method to represent LiDAR PCD to reduce data dimension i.e. improve computational efficiency. The \emph{feature extractor} module, which is a Convolutional Neural Network (CNN), cooperates with the \emph{activity recognizer} to recognize \emph{person tracker} provided multiple-humans' activities, and simultaneously, minimize the KL divergence of variational autoencoder-based \emph{domain adaptation} module to diminish environment/subject-independent divergence. Our framework not only improves the performance of MPT-HAR state-of-arts in supervised learning scenario, but also the domain adaptation techniques significantly improve HAR performance to facilitate low or unlabelled target environments/domains without losing any significant performance reduction in the edge computing system. The experimental results demonstrate the superiority of {\it PALMAR} frameworks and systems in terms of effectiveness and generalizability.
\begin{center}
\begin{figure}[!h]
  \includegraphics[ height = 1.75 in]{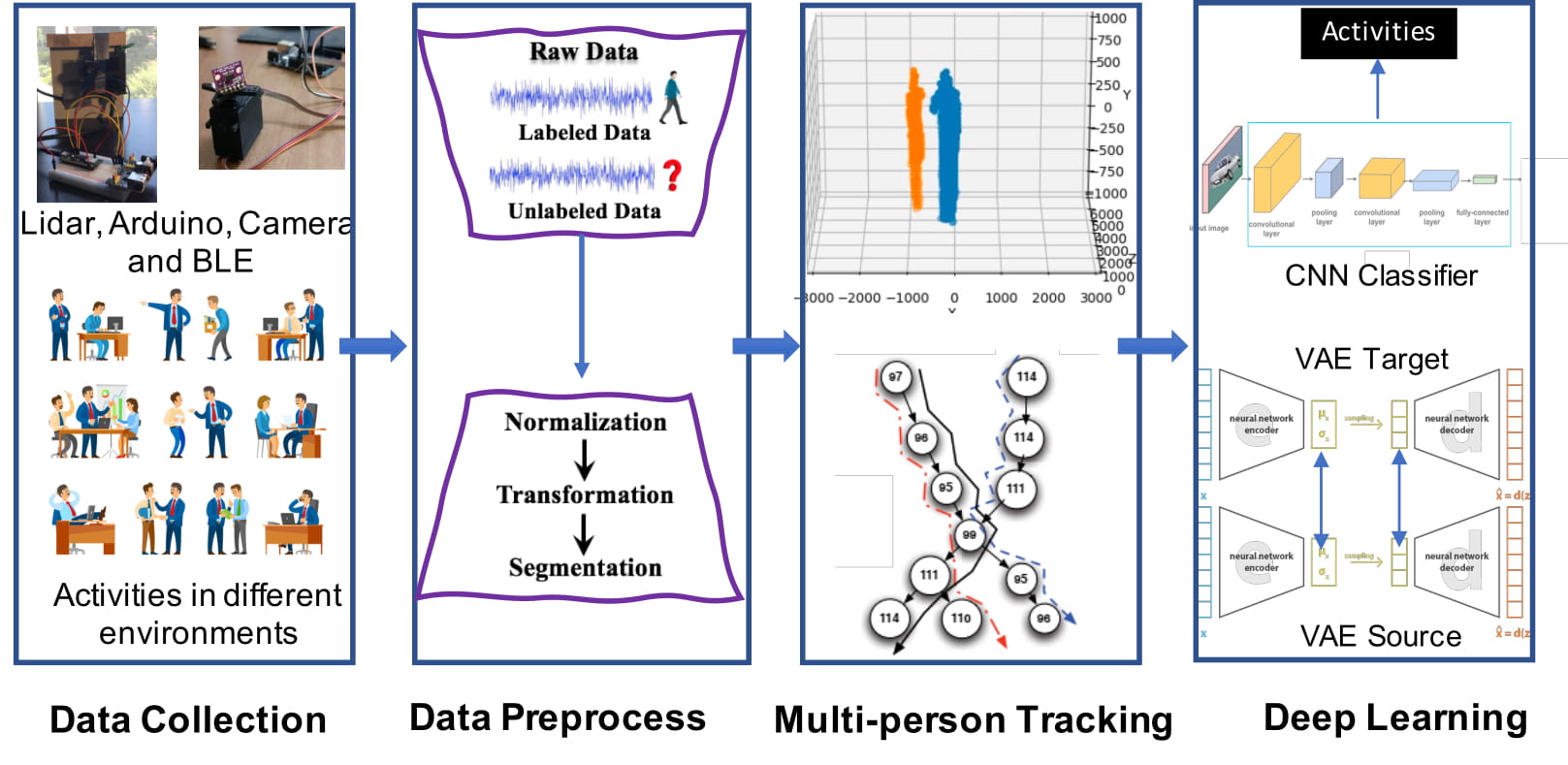}
  \caption{System Overview}
  \label{fig:overview}
\end{figure}
\end{center}
\section{System Overview}
As shown in Fig. \ref{fig:overview}, {\it PALMAR} consists of five components
\begin{itemize}
    \item {\bf Edge Computing Hardware Setup}: We integrated a $NVIDIA^R\;Jetson\;Nano^{TM}$ Developer Kit with the three PCD generating sensors (3D LiDAR and 79 GHz mmWave) and a camera using microUSB cable which has been used as our edge computing device in collecting data and system evaluation.
    \item {\bf Data Collection}: In this module, we consider a scenario where the human activities are monitored in different environments such as different sized rooms, indoor, outdoor, single inhabitant, multiple inhabitants. Our system first collects the activity data in each environment during the monitoring process. In this regard, we placed all of the testbed sensors in a cardboard along with an IP camera. The camera records the videos of the performed activities which have been used to label activity ground truth as well as computer vision-based HAR recognition.
    \item {\bf Data Preprocessing}: In this module, we first normalize the acquired signal and then transform the signal to a form suitable for analysis. Finally we split the transformed signal into short segments to train the activity recognition model. Then we represent the PCD to voxel format and apply signal processing techniques to fine tune PCD.
    \item {\bf Multiple person tracker}: In this module, we apply BIRCH and DBSCAN clustering method to identify number of persons present in the field of view (FOV) and track multiple persons related PCD clusters centroid using an Adaptive Order Hidden Markov (AO-HMM) Model and a Crossover Path Disambiguation Algorithm (CPDA) algorithms.
    \item {\bf Deep Learning Model}: We develop a baseline CNN model that takes the pre-processed and multiple-person related voxelized PCD representation as input and recognizes activities of in real-time. We also proposed a deep variational autoencoder based domain adaptation model which incorporates a customized KL-divergence optimization technique to diminish the divergence between source and target models' bottlenecks and apparently improves the domain adaptation. This model can take advantages of computer vision HAR dataset to scale the activity recognition accuracy significantly via domain adaptation in presence of scarce target data.
\end{itemize}

\section{Multiple Person Tracker}

\begin{figure*}[!h]
  \includegraphics[height = 2.2 in]{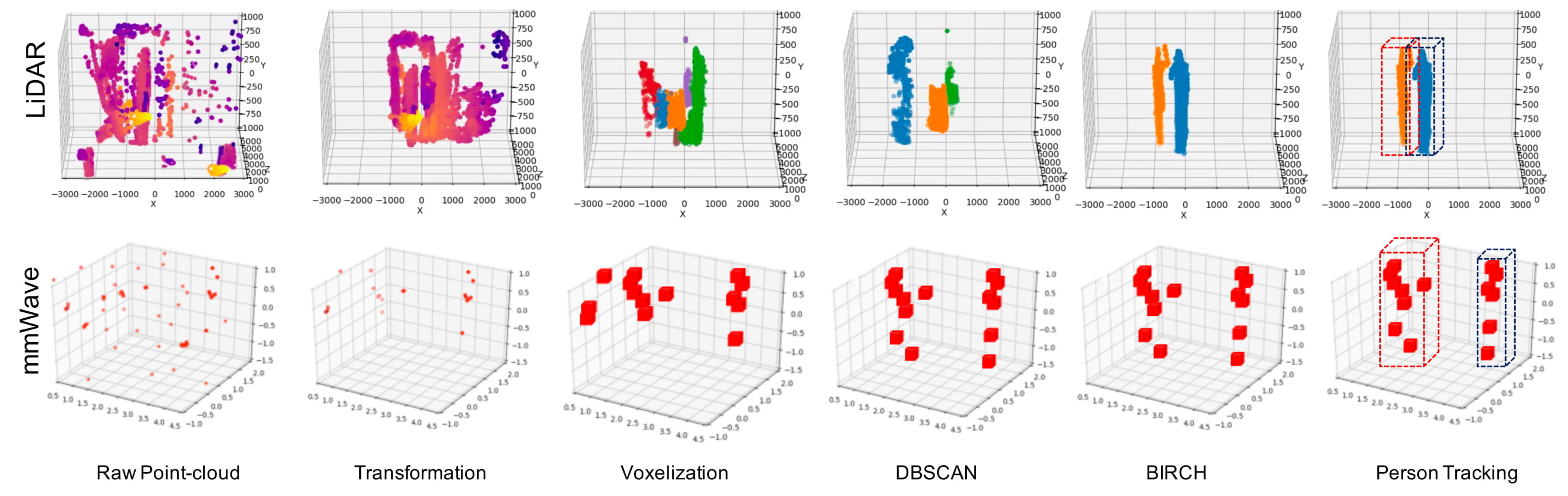}
  \caption{LiDAR (top) and mmWave (bottom) Point-cloud Processing Steps Visualization}
  \label{fig:lidar_point_cloud}
\end{figure*}

The person tracker framework consists of four modules that operate in a pipeline fashion as shown in Fig~\ref{fig:lidar_point_cloud}.

\subsection{Data Transformation}
At first, we convert the PCD sensors provided distance measure based spherical coordinates to Cartesian coordinates in our {\it transformation} phase of preprocessing. Then, we record the background with static room structure and subtract the background. Due to measurement variations there are quite a few points with near zero but, not exactly zero values. As part of transformation, we subtract minimal distance and maximum distance factors which has been determined using distance factor transformation method \cite{alam20}.

\subsection{Voxel Fitting}
Voxel fitting algorithm is popular in brain MRI processing, 3D scene analysis and depth camera analysis. We apply voxel fitting algorithm on PCD \cite{lopez05, hosoi13}. We run a series of experiments to determine right voxel fitting parameters i.e., length, breadth, and thickness of the voxel. In this regard, at first, we capture 3D grid of point density and variation profile of 4 different objects (laptop box, smart phone box, flower vast and printer) were created along a different axis. We measured the volume of each of the object and calibrated the corresponding 3D PCD distance ratio towards measuring accurate volume (length and height) of each object from three different distance measure, 1m, 2.5m and 5m. Apart from that, we also consider three different statistical voxel mapping methods: average mapping, average depth mapping (ADM) and average exclusive or mapping (AXOR). Then we run experiments on obtaining highest voxel resolution size with lowest error in detecting object volumes using allometric equation for object volume of estimation method  \cite{lopez05}.

\subsection{Clustering}
The generated voxel PCD are dispersed and not informative enough to detect distinct objects. Moreover, although static objects are discarded through our transformation phase (can be considered as background), the remaining points are not necessarily all reflected by moving people. To identify point-clouds generated from humans only, {\bf {\it PALMAR}} uses a sequence of clustering algorithms that gives more distinction between different objects in the frame. 

\noindent {\bf DBSCAN}: We use DBSCAN (density-aware clustering technique) to separate point-clouds of same cluster (i.e. same person) in 3D voxel space. A major advantage is that it does not require the number of clusters to be specified a priori, as in our case people walk in and fade out of the monitored scene at random. Additionally, DBScan can automatically mark outliers to cope with noise. However, in a real-world measurement study, we observed that points of the same person are coherent in the horizontal (x-y) plane, but more scattered and difficult to merge along the vertical (z) axis. We hence modify the Euclidean distance to place less weight on the contribution from the vertical z-axis in clustering:
\begin{equation}
D(p^i,p^j) = {(p^i_x - p^j_x)}^2 + {(p^i_y - p^j_y)}^2 +\alpha {(p^i_z - p^j_z)}^2
\end{equation}
where $p_i$ and $p_j$ are two different points and the parameter $\alpha$ regulates the contribution of vertical distance. Additionally, we perform fine tuning on all of the distance weights based on sample multi-person data sequences. Doing so, we place weights along the horizontal ($x$) and depth ($z$) distances. DBscan clustering thus will give back all relevant PCD and can be easily applied for multi-person tracking. Finally, we perform frame trimming to extract the final sequence for activity recognition training. During the DBscan clustering step, we elect to drop frames where there was no cluster identified nor a large enough cluster to constitute a person.

\noindent {\bf BIRCH}: Although, DBSCAN is a highly accurate in detecting number of clusters i.e. number of people in a given LiDAR PCD, it often fails when the number of people is large (say >2). DBScan also fails to categorize each data point in case of high dimensional sequence of data. Considering these limitations, we use BIRCH clustering algorithm after running DBSCAN clustering on voxelized PCD. The BIRCH (Balanced Iterative Reducing and Clustering using Hierarchies) algorithm is more suitable for the case where the amount of data is large and the number of categories $K$ is relatively large \cite{zhang96}. It runs very fast, and it only needs a single pass to scan the data set for clustering that empower BIRCH algorithm to more accurately track multiple clusters in a sequence of PCD. Optionally, the algorithm can make further scans through the data to improve the clustering quality. We use the power of BIRCH algorithm on DBSCAN provided clustered PCD and number of categories $K$. The BIRCH clustering algorithm consists of two main phases or steps (i) {\it Build the CF Tree}: In this phase, it loads the data into memory by building a cluster-feature tree (CF tree) and optionally, condenses this initial CF tree into a smaller CF. (ii) {\it Global Clustering}: In this phase, it applies an existing clustering algorithm (say DBSCAN) on the leaves of the CF tree and optionally, refines these clusters. Although, we are using a sequence of two clustering algorithms to fine-tune and cluster multiple person related PCD, due to the light weight nature of DBSCAN and BIRCH methods, it does not cost significant time while detecting clusters in real-time.

\subsection{Person Tracking}
To capture continuous individual PCD to track and identify a person, we require an effective temporal association of detections as well as correction and prediction of sensor noise. In this regard, we consider each cluster centroid (after DBSCAN and BIRCH ) as nodes in the 3D voxel grid that formulate a problem of multiple-particle tracking in 3D space. However, tracking the centroid using sequential multiple particle filter models will face the following challenges
\begin{enumerate}
    \item Requirement of fast tracking of individual targets i.e. cluster centroid from a static voxelized PCD network in the infrastructure. This needs to resolve unreliable node sequences, system noise and path ambiguity.
    \item Requirement of scaling for multi-user tracking where user motion trajectories may crossover with each other in all possible ways.
    \item After multi-user tracking of cluster centroids, re-cluster voxelized PCD for activity recognition of each person.
\end{enumerate}

We propose to use Adaptive Order Hidden Markov Model (AO-HMM) and Crossover Path Disambiguation Algorithm (CPDA) to address the above challenges \cite{de12}.

{\bf Adaptive Order Hidden Markov Model (AO-HMM)}: AO-HMM is a modified Hidden Markov Model (HMM) with a discrete time stochastic process. During state selection, AO-HMM chooses only the subset of states that are active and the neighbor (1-hop or 2-hop in Extended Activity Transition Graph) states i.e. order of HMM will be changed based on the number of active states and their neighbors. This reduces the computational complexity without compromising the accuracy of particle/cluster centroid tracking. However, The sub-state selection in AO-HMM also does not affect the optimality of HMM model and Viterbi computation in our application scenario as we apply AO-HMM on pre-constructed voxel space. Standard Viterbi decoding algorithm is modified for i) multiple observation, ii) multiple sequence decoding, and iii) fitting for activity awareness. For nonoverlapping motion, viterbi algorithm is computed on first order HMM \cite{rabi89} where transitions from time $(t-1)$ to $t$ are considered. For overlapping motion, viterbi algorithm is computed on second order HMM \cite{thede99} where transitions from time $(t-2)$ and $(t-1)$ to $t$ are considered.

{\bf Crossover Path Disambiguation Algorithm (CPDA)}: 
There were some constraints to directly using standard HMM model and Viterbi algorithm to our real-time application scenario. Regarding length of time window (say $W$) the standard Viterbi algorithm requires $O(W)$ operations. But the standard algorithm is not applicable in the case of a streamed input (with potentially no ending in sequence) and requirement of output within bounded delay. Regarding size of state space (say $S$), the standard Viterbi algorithm requires $O(S)^{2}$ operations, and still even on average $O(S\sqrt{S})$ operations by a modified version of Viterbi \cite{patel99}. On the other hand, the output state sequences from AO-HMM in each time window is partially disambiguated from the effect of path overlap or crossover. But it cannot always remove longer term path ambiguity that spreads beyond the Adaptive-HMM time window. To alleviate this, {\bf {\it PALMAR}} applies Crossover Path Disambiguation Algorithm or CPDA to the joint Adaptive-HMM output of last $C$ number of time windows. Unlike FindingHumo \cite{de12} where authors addressed binary motion sensor-assisted smart home motion tracking of multiple persons in 2D space, our problem domain lies on 3D space i.e. multi-person's path ambiguity, node sequences and trajectory crossover spread over all x, y and z axis. In this regard, we reduce the dimension to 2D by removing z-axis of our 3D voxelized PCD system after clustering phase and apply AO-HMM and CPDA algorithms on it. While AO-HMM addresses the multiple cluster (i.e. multiple person) sequence tracking and CPDA handles multi-person's path ambiguity and trajectory crossover.

\begin{figure}[!h]
  \includegraphics[height = 2.5 in]{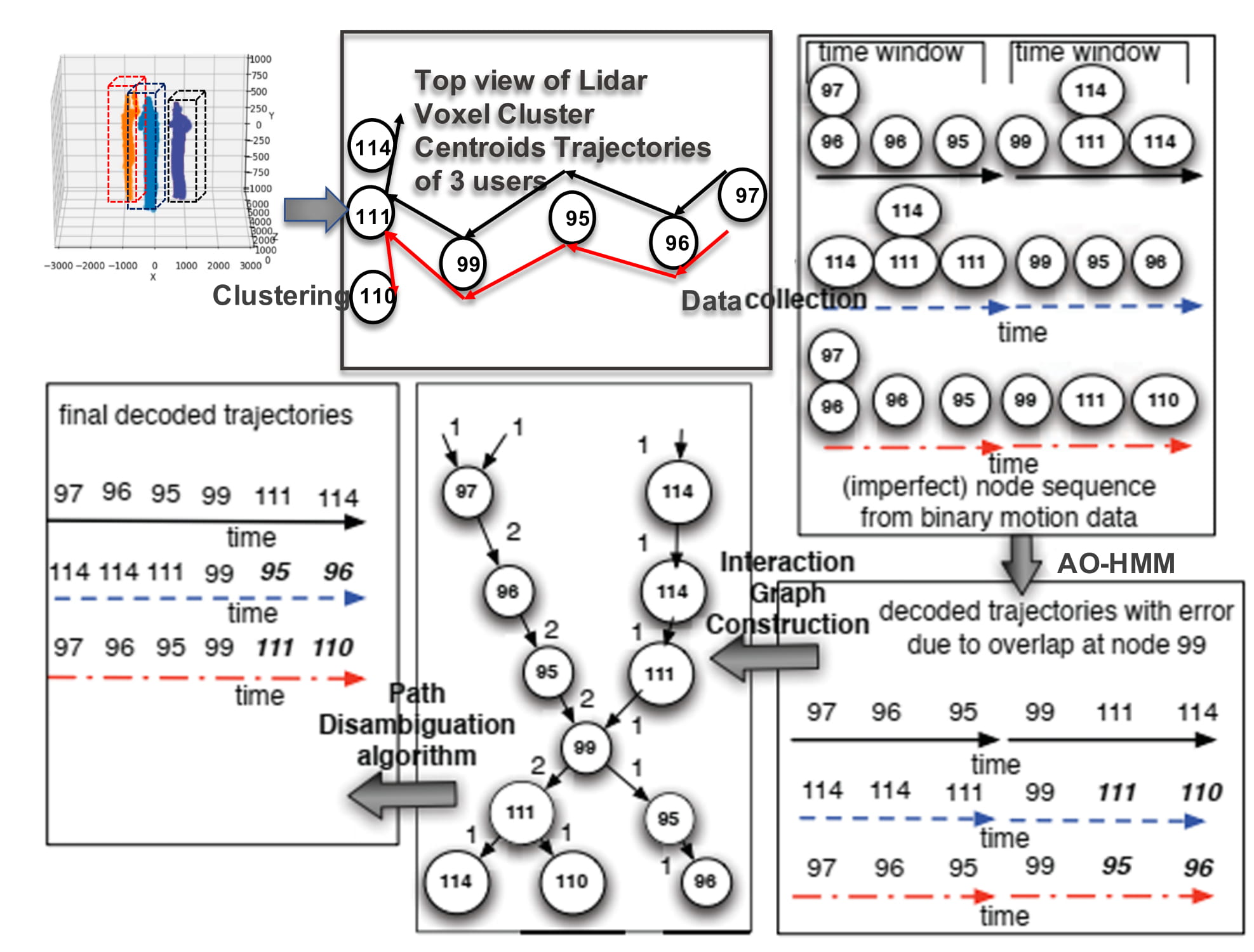}
  \caption{Working Example of 3 Inhabitants Person Tracking on Cluster Centroids (Nodes)}
  \label{fig:working_example}
\end{figure}

{\bf Working Example on Our Target Scenario}: Figure \ref{fig:working_example} illustrates how AO-HMM and CPDA algorithms together solve multiple-person tracking problem on LiDAR PCD in 3D voxel space. After applying several signal processing and clustering techniques, we have cluster centroid for each person in a multi-inhabitant smart home. We pass the cluster nodes to AO-HMM and later to CPDA algorithms. The unreliable cluster node sequence with system noise spread over 3D voxel space has been reduced first by applying dimension reduction to 2D space which is the input of AO-HMM algorithm. This is refined by applying AO-HMM in the next step. The decoded state sequence may still contain error due to path crossover (e.g. crossover of decoded path for user 2 and user 3 at node 99 in the Fig \ref{fig:working_example} ). This is further corrected by stitching the decoded paths and forming an Interaction Graph, which is then disambiguated by applying proposed CPDA algorithm. This results in final decoded motion trajectories. It is worth mentioning that the position of user is presented in form of sensor nodes’ position. Thus the tracking accuracy will be more (w.r.t the actual physical location of user) if the number of cluster is less (say <3).

\begin{figure*}[!htb]
\begin{minipage}{0.30\textwidth}
\begin{center}
\vspace{.5in}
  \includegraphics[width=1\linewidth]{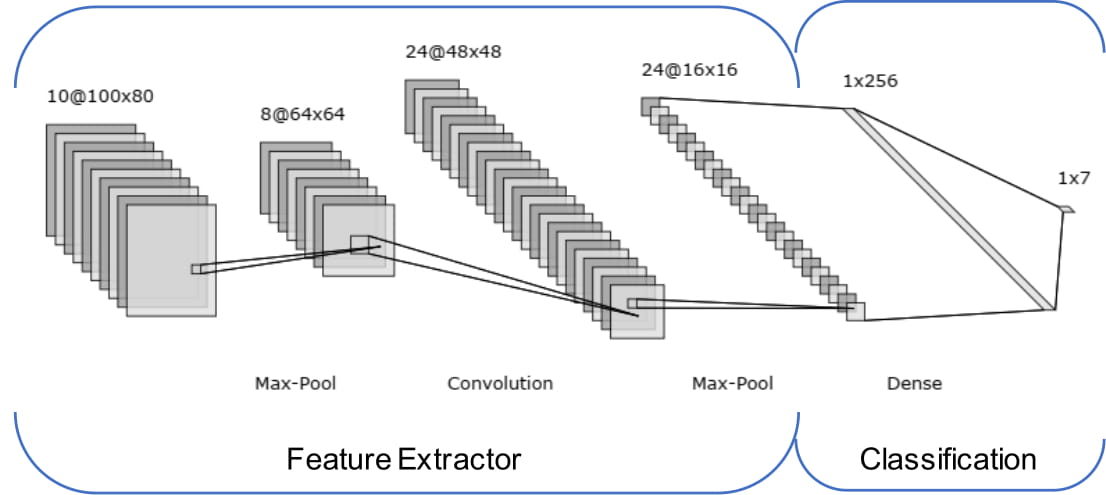}
  \vspace{.5in}
  \caption{Baseline Deep Activity Recognizer}
  \label{fig:recognizer}
\end{center}
\end{minipage}
\begin{minipage}{0.72\textwidth}
\begin{center}
  \centering
  \includegraphics[width=0.9\linewidth]{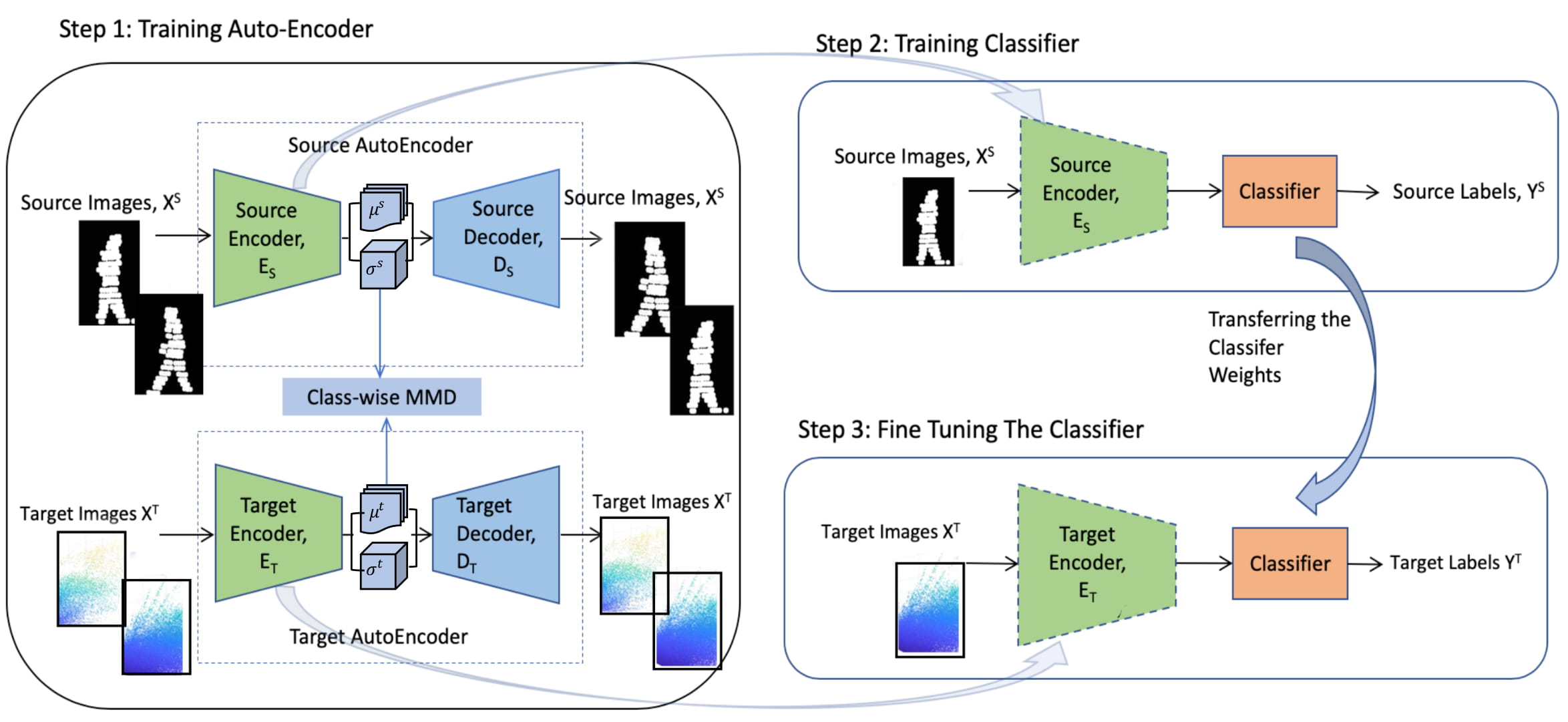}
  \caption{Our Variational Autoencoder Domain Adaptation Framework}
  \label{framework_arc}
\end{center}
\end{minipage}

\end{figure*}

\begin{figure}[!htb]
\begin{center}
  \centering
  \includegraphics[width=0.9\linewidth]{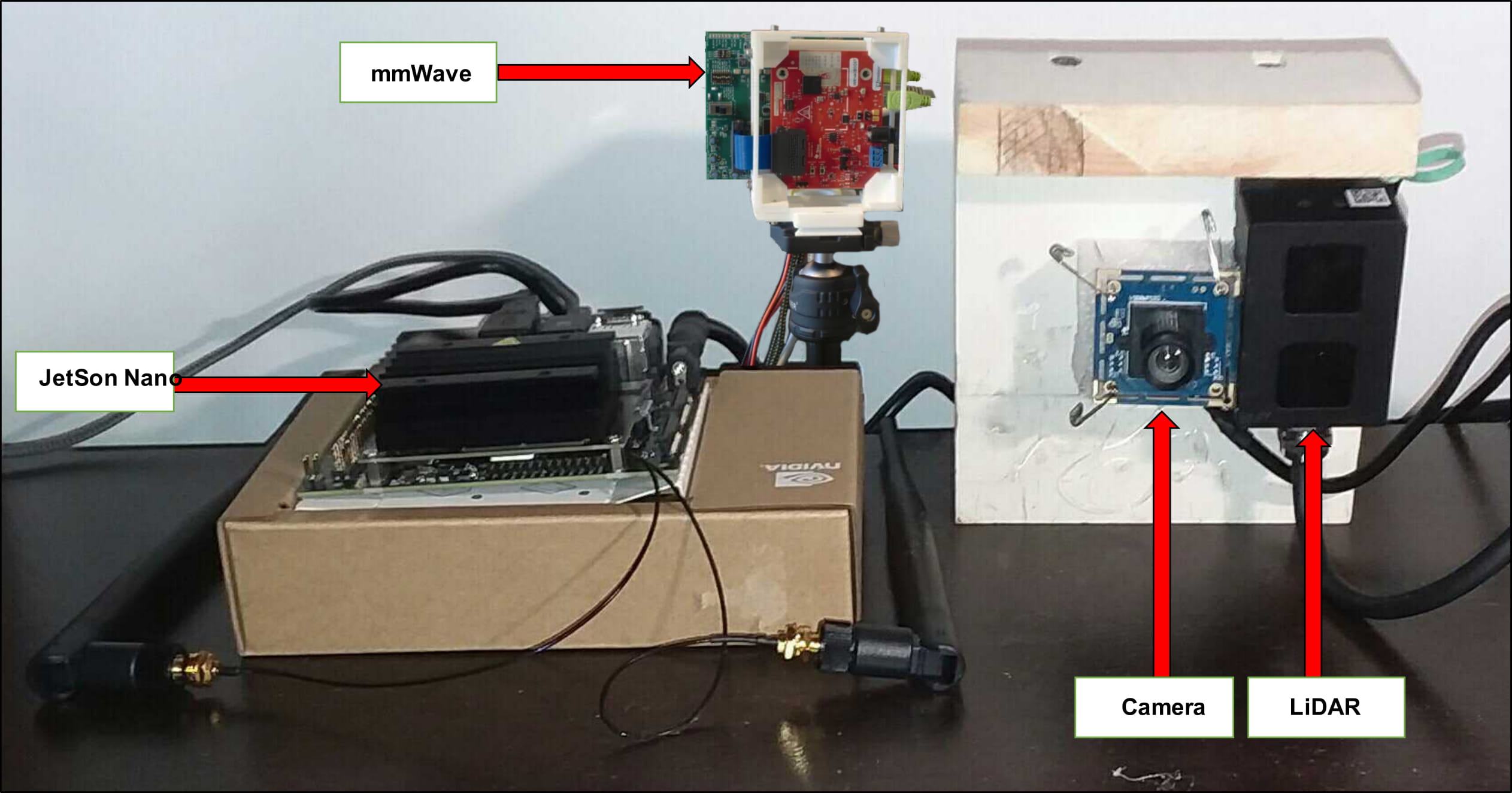}
  \caption{Our Hardware System}
  \label{fig:hardware}
\end{center}

\end{figure}

\section{Heterogeneous Domain Adaptation via Variational Autoencoder}
\subsection{Problem Formulation}
We can have our source domain $\mathcal{D}_s = \{(\mathbf{x}_i^s, \mathbf{y}_i^s)\}^{n_s}$, where $\mathbf{x}_i^s = \{x_1, . . . , x_{n_s}\} \in \mathbb{R}^{d_s}$ is the $i^{th}$ source data with $d_s$ feature dimension and $n_s$ sample size. Here, source labels $\mathbf{y}_i^s \in \mathcal{Y}_s$, where $\mathcal{Y}_s = \{1,2,...,n_c\}$ is the associated class label of corresponding $i^{th}$ sample of the source data $\mathbf{x}_i^s$. The source classification task $\mathcal{T}_s$ is to correctly predict the labels $\mathbf{y}^s$ from the data $\mathbf{x}^s$. The task $\mathcal{T}_s$ can also be viewed as a conditional probability distribution $P(\mathbf{y}_s|\mathbf{x}_s)$ from the probabilistic perspective. Similar to the source domain, now we let our target domain $\mathcal{D}_t = \mathcal{D}_l \cup \mathcal{D}_u = \{(\mathbf{x}_i^l, \mathbf{y}_i^l)\}^{n_l} \cup \{(\mathbf{x}_i^u\}^{n_u}$ where $\mathcal{D}_l$ and $\mathcal{D}_u$ is the labeled and un-labeled target data respectively. $\mathbf{x}_i^l \in \mathbb{R}^{d_t}$ is the $i^{th}$ labeled and $\mathbf{x}_i^u \in \mathbb{R}^{d_t}$ is the $i^{th}$ un-labeled target data with $d_t$ dimensional feature space. $ \mathbf{y}_i^l \in \mathcal{Y}_t$ is the corresponding class label of $i^{th}$ labeled target data. In our domain adaptation setting, $n_s \gg n_l$ and $n_u \gg n_l$. The target classification task, $\mathcal{T}_t$ is to predict the target class labels $\mathbf{y}^u_i$ with the unlabeled target data $\mathbf{x}^u_i$, more specifically, learning the probability distribution $P(\mathbf{y}^u_i|\mathbf{x}^u_i)$.

\par Assume that the source domain data $\mathbf{x}^s$ consists of source domain specific information $U^s$ and domain invariant information $V$. On the other hand, the target domain data $\mathbf{x}^t$ consists of target domain specific information $U^t$ and domain invariant information $V$. The goal of our work is to map the domain invariant information of both source and target domain into a common feature representation space which holds the domain invariant information $V$ of both domains. If we denote the $\mathbf{Z}$ symbol as the representation space, e.g. $\mathbf{Z}^s$ is the feature representation space of $\mathbf{x}^s$,
\begin{equation}
    P(\mathbf{Z}^s|\mathbf{x}^s) = P(\mathbf{Z}^t|\mathbf{x}^t) = V \\
\end{equation}

\subsection{Deep Activity Recognizer}
We develop a basic deep convolutional neural network (CNN) model as a baseline activity recognizer as shown in Fig \ref{fig:recognizer}. The baseline activity recognizer consists of two components: (i) feature extractor component consists of a CNN layer, a maxpooling layer, a CNN, a maxpooling, a fully connected; (ii) classification component consists of two fully connected layer with an output dimension of number of classes. We consider this baseline activity recognition structure as our base structure for variational autoencoder based domain adaptation model.

\subsection{Domain Adaptation with Variational Auto-Encoder}
As depicted in figure \ref{framework_arc}, our proposed framework have three main components, the Source variational auto-encoder, the target variational auto-encoder and the classifier. Each of the autoencoder has the structure of the feature extractor component of baseline deep activity recognizer. Our complete training process consists of two stages, training the variational auto-encoders and training the classifier. 

\subsubsection{Training Variational Auto-encoders}
In this training stage, source encoder $E_s$ and the target encoder $E_t$ map the input data $\mathbf{x}^s$ and $\mathbf{x}^t$ to the feature spaces $\mathbf{Z}^s$ and $\mathbf{Z}^t$ respectively in a probabilistic fashion. In our implementation of variational auto-encoder, we used gaussian function to sample the latent space $\mathbf{Z}$ from learned mean and varience matrix. The goal of the training is to minimize the distribution discrepancy between the source and target feature spaces $\mathbf{Z}^s$ and $\mathbf{Z}^t$. 
\begin{equation}
    \mathbb{Z}^s = E_s(\mathbf{x}^s) \sim \mathcal{N}(\mathbf{\mu}_s  (\textbf{x}^s), \mathbf{\sigma}_s(\textbf{x}^s))\\
\end{equation}

\begin{equation}
    \mathbb{Z}^t = E_t(\mathbf{x}^t) \sim \mathcal{N}(\mathbf{\mu}_t  (\textbf{x}^t), \mathbf{\sigma}_t(\textbf{x}^t))\\
\end{equation}

We use KLD loss function to minimize this discrepancy along the training. The source data $\textbf{x}^s$ and the target data $\textbf{x}^t$ are fed to the source and target auto-encoder respectively. Only the self reconstruction loss of variation auto-encoder is set to the source auto-encoder optimization loss added with KLD loss between the representation space $\mathbf{Z}^s$ and source reconstruction $\mathbf{x'}^s$.
\begin{eqnarray}
    \mathcal{L}_s = - \frac{1}{N} \sum_{i=1}^N x^s_i \cdot log(p(x^s_i)) \\\nonumber
    + (1-x^s_i) \cdot log(1-p(x^s_i) + \alpha \cdot \text{KLD}(Z^s_i, x'^s_i)
\end{eqnarray}
Where, $p(x^s_i)=E_s(D_s(x^s_i))$ is the prediction probability of $x^s_i$ by the source encoder and decoder and $\alpha$ is the weighting parameter for KLD loss.
On the other hand, the weighted sum of self reconstruction loss and the KLD loss function between the source and the target representation space is set to the target auto-encoder as loss function.
\begin{equation}
    \mathcal{L}_t = \mathcal{L}_{recon} + \beta \cdot \mathcal{L}_{KLD}\;
\end{equation}

\begin{equation}
    \text{where, } \mathcal{L}_{recon} =  - \frac{1}{N} \sum_{i=1}^N x^t_i \cdot log(p(x^t_i)) + (1-x^t_i)
\end{equation}
Here $\beta$ is the weighting parameter in order enforce the comparative importance of the reconstruction loss and KLD loss.

The Kullback-Leibler divergence (KLD) \cite{van2014renyi} is a powerful metric for determining divergence between two marginal probability distributions. The main idea of KLD is to calculate the probability divergence between the source and the target distributions. So, the KLD loss function is defined as,
\begin{equation}
    \mathcal{L}_{KLD}(\textbf{x}^s, \textbf{x}^t) = \frac{1}{N} \sum_{i=1}^N \frac{p(\textbf{x}^s) \cdot log(p(\textbf{x}^s))}{p(\textbf{x}^t)}
\end{equation}
Where $p(\textbf{x}^s)$ and $p(\textbf{x}^t)$ are the probability distribution of $\textbf{x}^s$ and $\textbf{x}^t$ respectively.

\subsubsection{Training the Classifier}
After training the source and the target auto-encoders, we train a common classifier network with the learned feature space $\textbf{Z}^s$ of the source encoder and the labels of the source network. We then transfer the learned classifier to the target network and use to classify the target feature representation $\textbf{Z}^t$. During the learning of the classifier, we make the source encoder network frozen. So, the learning objective function in case of learning the classifier is,
\begin{equation}
    \min_{f_c} \mathcal{L}_c[\textbf{y}^s, f_c(\textbf{Z}^s)]
\end{equation}
Here, $f_c(\cdot)$ represents the classifier network. We use categorical cross-entropy loss for classifier network. So, the loss function,
\begin{equation}
    \mathcal{L}_c = - \sum_{i=0}^C y^s_i \cdot log(f_c(Z^s_i))
\end{equation}

\section{Experimental Evaluation}
In this section, we describe our hardware system development, experiments on real-world collected data and existing data, and evaluate the performance of our frameworks.
\subsection{Hardware System Development}
We used Hypersen 3D Solid-state LiDAR sensor with the model number HPS-3D160. The detector provided $160\times60$ resolution which translates to $9600$ points for each frame. The detector has a field of view (FOV) of $76^o \times 32^o$ and uses light of wavelength 850 nm in the near-infrared band. Each frame was composed of a $160 \times 60$ grid with each element representing a distance value. We integrated a TexasInstrument 79-81 GHz mmWave sensor that provides 1024 PCD on each frame. We integrated a $NVIDIA^R\;Jetson\;Nano^{TM}$ Developer Kit with the LiDAR using microUSB cable. Jetson Nano delivers the compute performance to run modern AI workloads at unprecedented size, power, and cost (~100 USD). It is also supported by NVIDIA JetPack, which includes a board support package (BSP), Linux OS, NVIDIA CUDA, cuDNN, and TensorRT software libraries for deep learning, computer vision, GPU computing, multimedia processing, and much more. We also integrated an 8 MegaPixel USB camera (Sony IMX179 Sensor WebCam) to record the events for ground truth labeling (Fig \ref{fig:hardware}). We wrote a python USB data reader to read LiDAR and mmWave data directly thus we can run our algorithms in real-time.

\subsection{Data Collection}

{\bf {\it PALMAR} Dataset}: We employed 6 volunteers (graduate, undergraduate and high school students) as subjects and collected $7$ different activities ("bending", "call", "check\_watch", "single\_wave", "two\_wave", "walking", "normal\_standing") in 3 different rooms as indoor use case and one outdoor use case. The methodology for each session was designed as follows: subject would enter frame, subject would perform activity, subject then walks out of frame. For multiple person tracking, we use the same above strategies but with a 5 seconds interval for each person, i.e., the first person entered the field of view and start performing his/her assigned activity. After 2 seconds of first person's entrance to the field of view, second person entered and started performing activities. We followed the same strategy for the third person as well. As the first person started his/her activity first, he/she finished his/her task and left. Then the second person and finally the third person left the field of view. We willingly created 20 crossover events (both of the persons are in the same line of the LiDAR field of view) to create ambiguity of tracking to validate the performance of our crossover ambiguity algorithm. In total we have 45 minutes of data where 25 minutes of data belong to single inhabitant and 20 minutes of data belong to multiple-inhabitants. W used camera recorded videos as well as as the LiDAR PCD visualization for activity ground truth labeling. We define the entire LiDAR field of view as 1000 x 800 resolution and mmWave field of view as 300 x 200 of voxel spaces. While annotating ground truths, we label the X-Y plane for location annotation where we consider the head location of each person as centroid ground truth of multi-person tracker.

{\bf Benedek Dataset}: To compare our framework's efficacy, we also included existing large-scaled high resolution Velodyne HDL 64-E sensor data \cite{benedek18}. The dataset includes 28 participants who performed $5$ different activities ("bending", "call", "check\_watch", "single\_wave", "two\_wave") in 6 different rooms as indoor use case and a thorough study on hundreds of pedestrians as outdoor use case.

\subsection{Baseline Methods}
We implemented the following baseline methods to compare our multi-person tracking algorithm's efficiency:
\begin{itemize}
    \item {\bf Tracker 1: mID Method}: mID \cite{zhao19} proposed Hungarian Algorithm assisted Kalman Filter tracking solution to track multiple cluster centroids produced by DBSCAN algorithm on mmWave generated PCD. The central differences between mID and {\bf {\it PALMAR}} provided multiple person tracking solution are, (i) instead of using only DBSCAN, we used BIRCH clustering algorithm where we use DBSCAN algorithm only on each leaf node clustering, (ii) apart from the clustering, we incorporated an Adaptive Order HMM model to smooth the person tracking and CPDA algorithm to disambiguate during crossover trajectories among inhabitants.
    \item {\bf Tracker 2: {\it PALMAR} tracker without BIRCH}: In this method, we incorporated every framework we developed except BIRCH clustering algorithm.

    \item {\bf Tracker 3: {\it PALMAR} tracker without AO-HMM}: In this method, we incorporated every framework we developed except Adaptive Order HMM algorithm based smoothing technique of person tracking.

    \item {\bf Tracker 4: {\it PALMAR} tracker without CPDA}: In this method, we incorporated every framework we developed except CPDA based disambiguation technique of person crossover.
\end{itemize}

To compare the performance improvements of our proposed domain adaptation model, we implemented 7 stste-of-art algorithms: 1) DANN (Domain-Adversarial Training of Neural Networks)  \cite{gain16}; 2) CORAL \cite{sun16}; 3) ADR (Adversarial Dropout Regularization) \cite{saito18}; 4) VADA: A Virtual Adversarial Domain Adaptation (VADA) \cite{rui18}; 5) DIRT-T: Decision-boundary Iterative Refinement Training with a Teacher \cite{rui17}; 6) ADA: Associative Domain Adaptation \cite{philip17}; and 7) SEVDA: Self-ensembling for visual domain adaptation \cite{french18} models.

\subsection{Point-Cloud Processing for Each Sensor}
For each of the sensor PCD, we ran the data transformation and voxel fitting. For 3D LiDAR, we found optimized voxel grid resolution of 0.3 cm and voxel mapping method AXOR mapping that provided highest accuracy on each object volume estimation (average error rate 0.03m) using LiDAR PCD from all three considered distances which reduces the number of PCD significantly with maximum 90\% and minimum 79\% of PCD. For 79 GHz mmWave, we found voxel grid resolution 0.5 cm, voxel mapping method AXOR mapping with verage error rate 0.045m that reduces the number of PCD significantly with maximum 95\% and minimum 83\% of PCD.
 
\subsection{Results}
\subsubsection{Person Tracker Performance}
Since, Benedek Dataset does not have any ground truth of multiple person tracking, we evaluated the performance of our proposed multiple person tracking framework only on {\it PALMAR} dataset. At first, we ran the PCD transformation, voxelization and clustering (DBSCAN and BIRCH) algorithms to identify the centroid of each cluster i.e. the centroid of each inhabitant of the field of view. Then, we train the AO-HMM and CPDA algorithms using the location ground truth of each person. We use the Euclidean Distance (ED) as an error metric between the detected and original person location to evaluate person tracker's performance. Our person tracker provided an overall ED of 9.3 with a minimum ED as 2.5 and a maximum ED of 11.6 in tracking multiple persons in different scenarios for 3D LiDAR; and an overall ED of 11.5 with a minimum ED as 3.1 and a maximum ED of 13.6 in tracking multiple persons in different scenarios for mmWave. Table \ref{tab:person_tracker_performance} shows the details of the above results. We can clearly see that our person tracker outperforms the existing state-of-art model (Tracker 1: mID Method \cite{zhao19}) for every case with an overall 63.5\% of accuracy improvements for LiDAR and an overall 57.4\% of accuracy improvements for mmWave. The table also clearly illustrated the importance of each of our proposed modules as tracker 2, tracker 3 and tracker 4 individually improves the state-of-art method significantly for single, two and three inhabitants environment as well as outdoor scenario for both LiDAR and mmWave sensors. In case of crossover sessions, where we willingly created some ambiguity of crossing over the field of views, we can clearly see that without our proposed AO-HMM (Tracker 3) or CPDA (Tracker 4), person tracker significantly failed with ED. State-of-art multiple person tracker method mID (Tracker 1) also failed significantly in presence of crossover ambiguity. However, inclusion of AO-HMM and CPDA improved the performance significantly with improvements of 66.6\% and 51.2\% for LiDAR and mmWave sensors. On the other hand, in outdoor scenario, our proposed person tracker performed significantly better (55.5\% and 63.6\% for LiDAR and mmWave improvements) than state-of-art person tracking method as well.

\begin{table}[h!]
  \begin{center}
    \caption{Euclidean Distances between person trackers provided centroid and ground truth centroid of each person}
    \label{tab:person_tracker_performance}
    \begin{tabular}{ |p{0.8cm} p{.9cm}| p{.8cm}| p{.8cm}| p{.8cm}| p{.8cm}| p{.6cm}|}
\hline
  & & Tracker 1 & Tracker 2 & Tracker 3 & Tracker 4 & Ours\\ 
 \hline
 Single &LiDAR& 15.3 & 14.2 & 11.7 & 7.3 & 2.5 \\ 
  Person&mmWave& 16.6 & 16.1 & 15.3 & 9.5 & 6.9 \\ 
 \hline
 Two &LiDAR& 21.8 & 19.3 & 17.4 & 14.7 & 9.4 \\  
 Persons&mmWave& 26.7 & 25.3 & 23.9 & 16.3 & 12.5 \\  
 \hline
 Three&LiDAR& 30.2 & 23.7 & 17.3 & 14.9 & 10.2 \\  
 Persons&mmWave& 37.8 & 33.5 & 30.6 & 25.5 & 21.3 \\  
 \hline
  Crossover &LiDAR& 34.8 & 25.5 & 30.3 & 34.2 & 11.7 \\  
  Sessions&mmWave& 39.6 & 36.4 & 35.2 & 27.0 & 19.3 \\  
 \hline
   Outdoor &LiDAR& 23.6 & 24.9 & 17.5 & 14.3 & 10.5 \\  
  Sessions&mmWave& 28.5 & 27.4 & 22.3 & 16.6 & 10.3 \\  
 \hline
   Overall &LiDAR& 25.7 & 21.2 & 16.8 & 12.3 & 9.3 \\  
   ED&mmWave& 27.5 &  24.6 & 20.4 & 17.5 & 11.6 \\  
 \hline
    \end{tabular}
  \end{center}
\end{table}

\subsubsection{Activity Recognition Results}
After identifying cluster's centroids (i.e. each person's location) and extracting their related PCD (as stated in Section III.C), we ran the activity recognition algorithm on each of the person's extracted PCD voxels for both {\it PALMAR} datasets and Benedek dataset. We converted the our LiDAR, mmWave and Benedek datasets as 1000 x 800, 300 x 200 and 3500 x 2000 resolution (as used in Benedek et. al. \cite{benedek18} experiments) voxels. Then we trained and test on our Deep Activity Recognizer model on both of the datasets. We achieved accuracies of  $75.75\% \pm0.1$, $70.77\% \pm0.2$ and $73.85\% \pm0.1$ for LiDAR, mmWave and BENEDEK datasets source only respectively. We also achieved an accuracy of $93.81\% \pm0.1$ using only camera videos data (Computer Vision).

\begin{table*}[!h]
  \caption{Activity recognition performances (\% Accuracy on target) of domain adaptation}
  \label{tab:result-table}
  \centering
  \begin{tabular}{|p{1.5cm}|p{1.4cm}|p{1.3cm}|p{1.5cm}|p{1.5cm}|p{1.3cm}|p{1.3cm}|p{1.3cm}|p{1.3cm}|p{1.3cm}|}
      \hline
    Method     & BENEDEK $\xrightarrow{}$ LiDAR  & LiDAR $\xrightarrow{}$ BENEDEK &
BENEDEK $\xrightarrow{}$ mmWave  & mmWave $\xrightarrow{}$ BENEDEK &
LiDAR $\xrightarrow{}$ mmWave  & mmWave $\xrightarrow{}$ LiDAR &
Video $\xrightarrow{}$ LiDAR  & Video $\xrightarrow{}$ mmWave &
Video $\xrightarrow{}$ BENEDEK 
    \\
    \hline
    Source & $73.85\pm.1$  & $75.75\pm.1$ & $73.85\pm.1$ & $70.77\pm.2$& $75.75\pm.1$& $70.77\pm.2$ & $93.81 \pm.1$& $93.81 \pm.1$& $93.81 \pm.1$\\    \hline
    DANN \cite{gain16} & 
    $82.83\pm.3$  & 
    $83.89\pm.1$ &
    $83.88\pm.1$ & 
    $85.34\pm.2$& 
    $82.43\pm.3$ & 
    $89.65\pm.2$& 
    $89.66\pm.3$ & 
    $91.56\pm.2$& 
    $90.34\pm.2$\\    \hline
    CORAL   \cite{sun16} & 
    $86.03\pm.3$ & 
    $84.57\pm.2$&
    $81.70\pm.1$ & 
    $83.63\pm.2$& 
    $81.75\pm.3$ & 
    $86.39\pm.2$& 
    $87.44\pm.3$ & 
    $88.60\pm.2$& 
    $89.56\pm.2$\\    \hline
    ADR  \cite{saito18}  & 
    $89.03\pm.4$ & 
    $84.66\pm.4$ &
    $85.92\pm.1$ & 
    $79.66\pm.2$& 
    $80.80\pm.3$ & 
    $87.32\pm.2$& 
    $86.3\pm.3$ & 
    $87.59\pm.2$& 
    $87.53\pm.2$\\    \hline
    VADA \cite{rui18}  & 
    $91.51\pm.2$ & 
    $85.05\pm.2$ &
    $87.44\pm.1$ & 
    $82.62\pm.2$& 
    $79.51\pm.3$ & 
    $84.98\pm.2$& 
    $87.21\pm.3$ & 
    $88.45\pm.2$& 
    $87.61\pm.2$\\    \hline
    DIRT \cite{rui17}  & 
    $92.65\pm.2$ & 
    $85.43\pm.2$ &
    $85.34\pm.1$ & 
    $86.98\pm.2$& 
    $83.58\pm.3$ & 
    $87.87\pm.2$& 
    $91.44\pm.3$ & 
    $91.56\pm.2$& 
    $91.56\pm.2$\\    \hline
    ADA \cite{philip17}  & 
    $92.78\pm.2$ & 
    $86.03\pm.2$ &
    $84.56\pm.1$ & 
    $88.50\pm.2$& 
    $85.78\pm.3$ & 
    $86.98\pm.2$& 
    $92.45\pm.3$ & 
    $92.56\pm.2$& 
    $90.89\pm.2$\\    \hline
    SEVDA \cite{french18} & 
    $92.13\pm.3$ & 
    $86.7\pm.3$  & 
    $83.58\pm.1$ & 
    $88.45\pm.2$& 
    $86.34\pm.3$ & 
    $87.86\pm.2$& 
    $93.56\pm.3$ & 
    $92.87\pm.2$& 
    $91.56\pm.2$\\    \hline
    \textbf{Ours} & 
    $\mathbf{94.63}\pm.3$ & 
    $\mathbf{89.38}\pm.2$& 
    $\mathbf{85.39}\pm.1$ & 
    $\mathbf{87.52}\pm.2$& 
    $\mathbf{90.64}\pm.3$ & 
    $\mathbf{91.88}\pm.2$& 
    $\mathbf{96.71}\pm.3$ & 
    $\mathbf{95.55}\pm.2$& 
    $\mathbf{94.38}\pm.2$\\    \hline
    
  \end{tabular}
\end{table*}

\begin{figure}[!htb]
\begin{minipage}{0.23\textwidth}
\begin{center}
   \epsfig{file=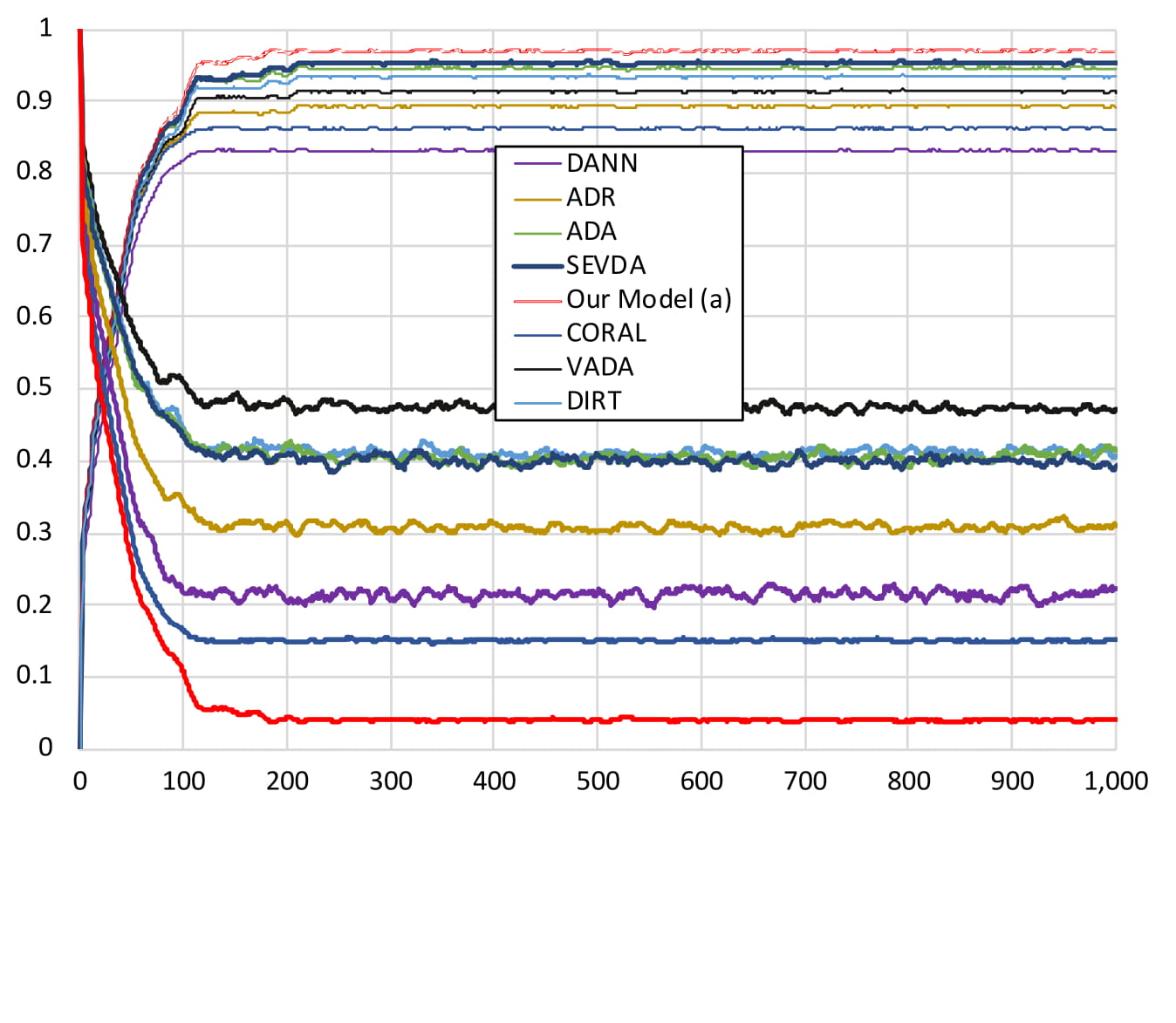,height=1.5in}
\caption{Video$\xrightarrow{}$LiDAR Domain Adaptation Accuracy Validation and Sensitivity}
   \label{fig:benedek_lamar}
\end{center}
\hfill 
\end{minipage}
\begin{minipage}{0.23\textwidth}
\begin{center}
   \epsfig{file=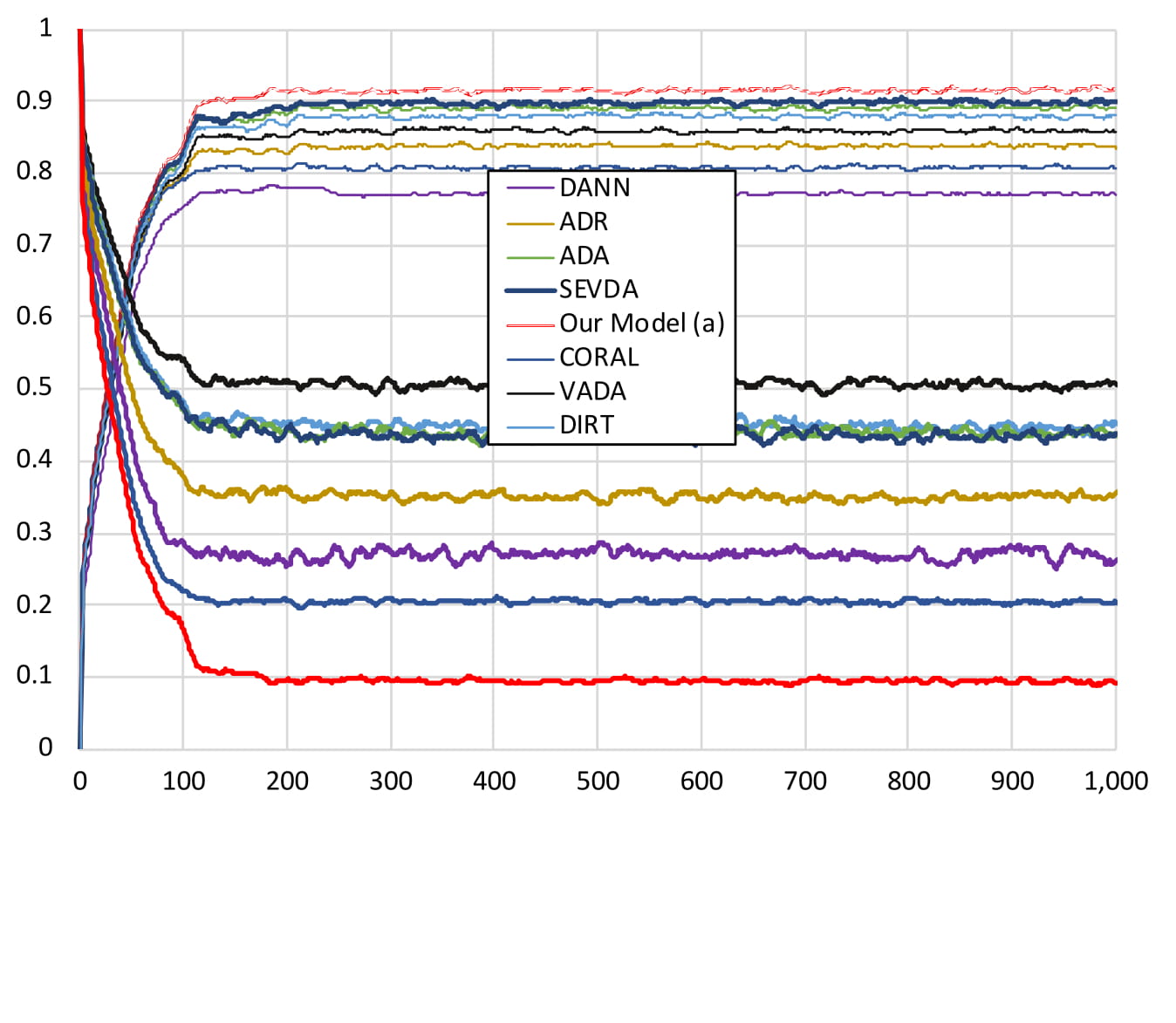,height=1.5in}
\caption{BENEDEK$\xrightarrow{}$LiDAR Domain Adaptation Accuracy Validation and Sensitivity}
   \label{fig:lamar_benedek}
\end{center}
\end{minipage}
\end{figure}

\subsubsection{Domain Adaptation Results}
The aim of our domain adaptation is to utilize more accurate models to improve less accurate models as stated above. In this regard, we experimented on all of our developed models which performances wise can be ranked as follows: $ComputerVision>LiDAR>BENEDEK>mmWave$. While selecting source and target dataset pair, we consider only the common classes between them to avoid class heterogeneity. According to the training method presented, we first trained the source autoencoder until convergence, froze the source encoder, trained the target autoencoder using the target labels by optimizing the Equation 8 and finally train the classifier network using the objective function presented in Equation 9. Table \ref{tab:result-table} presented the details results of domain adaptation using state-of-art methods as well as our proposed method. We clearly can see that our proposed framework outperforms all baseline domain adaptation methods for every pair of source-target datasets. From these figures  Fig \ref{fig:benedek_lamar} and Fig \ref{fig:lamar_benedek}, we clearly can identify that Video$\xrightarrow{}${\it LiDAR} domain adaptation converges much faster and less sensitive than {\it BENEDEK}$\xrightarrow{}$LiDAR. As we know, video data has lesser noises than LiDAR PCD data that depicts the faster convergence with lesser noises (sensitivity) for domain adaptation than {\it BENEDEK}$\xrightarrow{}$LiDAR. While trying to transfer knowledge from higher quality (Benedek) dataset to lower quality ({\it LiDAR}) data, we see significant improvements of accuracy as well as faster convergence of domain adaptation (Fig \ref{fig:benedek_lamar}). On the other hand, while trying to transfer knowledge from lower quality data ({\it LiDAR}) to higher quality data (Benedek), the improvement of domain adaptation is not that much significant and domain convergence takes more epochs than the prior one.

\begin{table}[h!]
  \begin{center}
    \caption{Time-delays (in seconds) of PALMAR system activity recognition with different baseline multi-person tracking models (+T1 means proposed domain adaptation framework plus Tracker 1)}
    \label{tab:system_performance}
    \begin{tabular}{ |c c| c| c| c| c| c|}
\hline
  & & +T1 & +T2 & +T3 & +T4 & +Ours\\ 
 \hline
 Single &LiDAR& 3.35 & 3.37 & 3.4 & 3.4 & 3.51 \\ 
  Person&mmWave& 2.67 & 2.69 & 2.67 & 2.67 & 2.78 \\ 
 \hline
 Two &LiDAR& 4.15 & 4.48 & 4.47& 4.40 & 4.55 \\  
 Persons&mmWave& 3.58 & 3.80 & 3.81 & 3.89 & 4.10 \\  
 \hline
 Three&LiDAR& 5.22 & 5.49 & 5.51 & 5.55 & 6.10 \\  
 Persons&mmWave& 4.39& 4.65 & 4.68 & 4.60 & 4.75 \\  
 \hline
   Overall &LiDAR& 3.95& 4.21 & 4.21 & 4.33 & 4.38 \\  
   Delay&mmWave& 3.51 &  3.79 & 3.80 & 3.84 & 4.01 \\  
 \hline
    \end{tabular}
  \end{center}
\end{table}

\subsubsection{Edge Computing Device Performance}
To evaluate the edge computing system performance, we ran {\it LAMAR} system in real-time crowd environment (hallway) and recorded videos along with the output of framework (number of person and each person's activities). We considered average time-delay (in seconds) of detected activity's start-end points comparing to the ground truth. Table \ref{tab:system_performance} shows that for single person scenario, the time delay is almost same to state-of-art (T1) but for multiple person scenario, our method produced higher time delays in recognizing activities with an overall time delay of 4 seconds which is $\approx 0.5$ seconds degradation of activity recognition time which is tolerable in terms of real-time HAR system.

\section{Related Works}
This paper builds on previous works on human activity recognition using machine learning, deep learning, domain adaptation, multiple inhabitant tracking and LiDAR PCD processing techniques. Here we compare and contrast our contributions with the most relevant existing literature.
\subsection{Multiple Person Tracking using Ambient Sensors}
Multiple person tracking has been a popular problem in computer vision where video streams have been processed to track by detection of human using supervised \cite{hamid11,andri12,huang08, alam19a,alam19b, alam17, alam16} or unsupervised methods \cite{ frag12,wang14}. However, tracking by detection is not applicable in case of LiDAR PCD where different body parts are not clearly visible by the light detection and ranging time of flight sensor like LiDAR. On the other hand, multiple densely implanted ambient sensors assisted multiple person tracking has also been proposed by many researchers \cite{de12, lei15} who proposed probabilistic path tracking of each sensor node firings using a pretrained model for supervised learning \cite{de12} or unknown number of inhabitants for unsupervised person tracking \cite{lei15}. On the other hand, using LiDAR or LiDAR like RF technologies (such as millimeter wave, RF sensor) for multiple person tracking is relatively new area of research  \cite{alvarez19, guer19, benedek18, zhao19}. Most of the PCD technologies (sensors that create PCD of object) utilize sensor generated high dimensional points-clouds in certain pipeline consists of PCD processing, representation, clustering and tracking. \cite{alvarez19} propsoed a CNN network based people's leg identification and tracking the leg motion using Kalman filter method, which requires the total view of human body work efficiently. \cite{guer19} proposed to use Kalman filter and LSTM (Long Short Term Memory) deep learning model to track multiple inhabitants in indoor scenario for Robots using 2D LiDAR, which fails in presence of furniture as well as in crossover ambiguity. \cite{benedek18} proposed multiple persons tracking using Kalman Filter and Gait Pattern to track pedestrians, however, it fails to address crossover ambiguity which is present in multiple inhabitant smart homes. \cite{zhao19} presented People Tracker package, aka PeTra, which uses a convolutional neural network to identify person legs in complex environments and develop a correlation technique to estimate temporal location of people using a Kalman filter, but, this method also fails in addressing crossover ambiguity.
\subsection{Domain Adaptation for Activity Recognition}
Among all of the domain adaptation techniques in activity recognition in computer vision, the most successful one is is the problem of cross-viewpoint (or viewpoint-invariant) action recognition \cite{24, 27, 31, 40, 46}. These works focus on adapting to the geometric transformations of a camera but do little to combat other shifts, like changes in environment such as indoor or outdoor. Works utilise supervisory signals such as skeleton or pose \cite{31} and corresponding frames from multiple viewpoints \cite{24, 46}. Recent works have used GRLs to create a view-invariant representation \cite{27}. Though several modalities (RGB, flow and depth) have been investigated, these were aligned and evaluated independently. On the other hand, before deep-learning, heterogenous domain adaptation (HDA) for action recognition used shallow models to align source and target distributions of handcrafted features \cite{4, 11, 64}. Three recent works attempted deep HDA \cite{7, 19, 36}. These apply GRL adversarial training to C3D \cite{54}, TRN \cite{63} or both \cite{36} architectures. Jamal et al.’s approach \cite{19} outperforms shallow methods that use subspace alignment. Chen et al. \cite{7} show that attending to the temporal dynamics of videos can improve alignment. Pan et al. \cite{36} use a crossdomain attention module, to avoid uninformative frames. Two of these works use RGB only \cite{7, 19} while \cite{36} reports results on RGB and Flow, however, modalities are aligned independently and only fused during inference. The approaches \cite{7, 19, 36} are evaluated on 5-7 pairs of domains from subsets of coarse-grained action recognition and gesture datasets, for example aligning UCF \cite{41} to Olympics \cite{34}. We evaluate on 6 pairs of domains. Compared to \cite{19}, we use 3.8x more training and 2x more testing videos. The EPIC-Kitchens \cite{8} dataset for fine-grained action recognition released two distinct test sets—one with seen and another with unseen/novel kitchens. In the 2019 challenges report, all participating entries exhibit a drop in action recognition accuracy of 12-20\% when testing their models on novel environments compared to seen environments.

\subsection{PCD based Activity Recognition}
LiDAR and mmWave based activity recognition has been explored by many researchers \cite{benedek18, wen18}. Benedek et. al. proposed to use DBSCAN clustering algorithm and Kalman Filtering method to identify number of people and track their movements \cite{benedek18}. Apart from that, \cite{benedek18} also proposed to utilize CNN based supervised method to detect multiple persons' 5 different activities. As, this is one of the rare investigated activity recognition framework using LiDAR, it does have many limitations that includes extremely poor accuracy of detecting activities (75\%) as well as person tracking (89\%). Wenjun et. al. proposed adversarial domain adaptation technique to detect HAR in presence of environmental diversity but did not cover the multiple inhabitant tracking problem \cite{wen18} . 

\section{Conclusion}
{\bf {\it PALMAR}} is the first of its kind adaptive activity recognition technique in multiple inhabitant point-cloud image generating technology-assisted environment with best accuracy ever achieved in indoor environment and outdoor environment. Although, we establish the state-of-art of PCD-based activity recognition, we have certain limitations. The data collection part was one of the most challenging part of this project due to the on-going COVID-19 pandemic related campus lock-down. To accommodate appropriate data collection, we recruited only our lab members by shipping the {\it PALMAR} system to their house. For multi-person activity data collection, participants were requested to engage their family members in home environment without exposing themselves to outdoor people. Due to these limitations, we were able to engage only 6 unique users with a maximum occupancy of 3 persons. However, we also could not collect significant amount of outdoor activity recognition data for multiple persons due to the on-going lock-down in the campus. In this project, we collected the gait information and  breathing rate (participants worn respiratory belt) of each participants. However, due to the limited number of users, we could not propose improved models for gait pattern based person identification/reidentification which we would significantly improve the accuracy of person tracking in case of multiple persons. In future, we aim to develop a breathing rate detection framework along with an experimental evaluation of {\it PALMAR} framework's time -complexity and real-time system performance. In terms of application, we aim to address multiple scenarios, such as older adults, dementia population and skilled-nursing facility inhabitants monitoring technique to accommodate more real-time solution for well-being of in need population. 

\section*{Acknowledgement}
We thank Fernando Mazzoni and Mohammad Haerinia for helping in data collection.

\end{document}